\documentclass[11pt,a4paper]{article}
\usepackage[hyperref]{acl2020}
\usepackage{times}
\usepackage{latexsym}

\usepackage{graphicx}
\usepackage{subfigure}
\usepackage{booktabs} 
\usepackage[none]{hyphenat}
\usepackage{bm,bbm,dsfont}
\usepackage{amsmath, amssymb}

\newcommand{\sts}{\textsc{Seq2Seq~}}

\DeclareMathOperator*{\argmax}{arg\,max}
\aclfinalcopy 

\title{Non-Autoregressive Neural Dialogue Generation}

\author{Qinghong Han$^{1*}$, Yuxian Meng$^{1*}$, Fei Wu$^2$ and Jiwei Li$^1$ \\
$^1$ ShannonAI \\
$^2$ Department of Computer Science and Technology, Zhejiang University\\
 {\{qinghong\_han, jiwei\_li\}}@shannonai.com, wufei@zju.edu.cn 
}

\date{}

\begin{document}
\maketitle
\begin{abstract}
Maximum Mutual information (MMI), which models the bidirectional dependency  between responses ($y$) and contexts ($x$), i.e., 
the forward probability $\log p(y|x)$ and the backward probability $\log p(x|y)$, 
has been widely used as the  objective
in the \sts model
 to address the  dull-response
 issue
  in open-domain dialog generation. 
Unfortunately, under the framework of the \sts model, 
direct decoding from $\log p(y|x) + \log p(x|y)$
is infeasible 
 since the
second part (i.e., $p(x|y)$) requires the completion of target generation before it can be  computed, and the
search space for $y$ is enormous. Empirically, an N-best list is first generated given $p(y|x)$, and $p(x|y)$ is then used to rerank the N-best list, which 
inevitably results
in non-globally-optimal solutions.

In this paper, we propose to use non-autoregressive (non-AR) generation model to address this  non-global optimality issue. 
Since 
target tokens are generated independently in non-AR generation, 
$p(x|y)$ for each target word can  be computed as soon as it's generated, and does not have to wait for the completion of the whole sequence.
This naturally resolves the non-global optimal issue in decoding. 
Experimental results demonstrate that the proposed non-AR strategy produces more diverse, coherent, and appropriate responses, yielding substantive gains in BLEU scores  and in human evaluations.\footnote{Qinghong and Yuxian contribute equally to this work.}
\end{abstract}

\section{Introduction}
\label{introduction}
Open-domain 
neural dialogue generation 
\citep{vinyals2015neural,sordoni2015neural,li2015diversity,mou2016sequence,serban2016multi,RE,mei2016coherent,serban2016hierarchical,serban2016building,serban2016,baheti2018generating,wang2018deep,ghazvininejad2018knowledge,zhang2018personalizing,gao2019neural} treats dialog contexts ($x$) as sources,and responses ($y$) as targets and uses the encoder-decoder model \cite{sutskever2014sequence,vaswani2017attention} as the backbone to generate responses. 
 \sts models offer the promise of scalability and language-independence, along with the
capacity to  capture contextual dependencies semantic and syntactic
relations between sources and targets.

One of key  issues with the \sts structure is that it exhibits a strong  tendency to generate dull, trivial or non-committal responses (e.g., \textit{I~don't know} or  \textit{I'm OK}) regardless of the input, which has been observed by many recent works \cite{li2015diversity,sordoni2015neural,serban2015hierarchical,niu2020avgout}.
Various strategies \cite{li2015diversity,vijayakumar2016diverse,baheti2018generating,niu2020avgout} have been proposed to address this issue, 
,  one of the most
widely used of which is to replace the MLE objective in the \sts training with the maximum mutual information objective (MMI for short)  \cite{li2015diversity}. 
MMI  models the bidirectional dependency  between responses ($y$) and contexts ($x$).
It takes the form of the linear combination of 
the forward probability $\log p(y|x)$ and the backward probability $\log p(x|y)$. 
The intuition behind MMI is straightforward: it is easy to predict a dull response given any context,  but hard to predict the context given a dull response since the context 
that corresponds 
to a dull response could be anything.

Unfortunately, under the framework of the \sts model, 
direct decoding from $\log p(y|x) + \log p(x|y)$
is infeasible 
 since the
second part (i.e., $p(x|y)$) requires the completion of target generation before $p(x|y)$ can be  computed, and the
search space for $y$ is huge. 
Empirically, an N-best list is first generated given $p(y|x)$, and $p(x|y)$ is then used to rerank the N-best list.
Due to the fact that 
beam search 
lacks for diversity in the beam: candidates often differ
only by punctuation or minor morphological variations, with most of the words overlapping,
this reranking strategy  
inevitably results
in non-globally-optimal solutions.
Some strategies have been proposed to alleviate this non-global-optimality issue, such as generating a more diverse N-best list  \cite{li2016simple,gu2017trainable,vijayakumar2016diverse},
or using reinforcement learning to estimate the future score of $p(x|y)$ \cite{li2017learning}, which help alleviate 
the non-globally-optimal issue, 
 but cannot fully address it. 

Non-autoregressive (non-AR) generation \cite{jiatao2018nat, xuezhe2019flowseq, lee2018iterative} provides resolution to the non-global-optimality issue. 
Under the formalization of non-AR generation, 
target tokens $y_t$ are generated independently, 
which enables 
 $p(x|y_t)$ to be computed 
  as soon as $y_t$ is generated.
This naturally resolves the non-global optimal issue in decoding. 
We conduct experiments on the widely used Opensubtitle dataset  and 
experimental results demonstrate that the proposed strategy produces more diverse, coherent, and appropriate responses, yielding substantive gains in BLEU scores  and in human evaluations.

The rest of this paper is organized as follows:
Section 2 and section 3 present related work and background knowledge respectively.
The propose model is described in Section 4. 
Experimental results and ablation studies are detailed in Section 5 and 6, followed by a brief conclusion in Section 7. 

\section{Related Work}
\subsection{Neural Dialogue Generation}
End-to-end neural approaches for dialogue generation 
use
 \sts architectures  \cite{sutskever2014sequence,vaswani2017attention} as the backbone to generate syntactically fluent and meaningful responses, providing the flexibility to capture contextual semantics between source contexts and target responses. Recent studies have endowed these models with the ability to  model contexts \cite{sordoni2015neural,serban2016hierarchical,serban2016building,tian-etal-2017-make,lewis-etal-2017-deal}, generating coherent and personalized responses \cite{li2016persona,zhao-etal-2017-learning,shao-etal-2017-generating,10.5555/3298023.3298055,zhang2018personalizing,bosselut-etal-2018-discourse}, generating uttterances with different attributes or topics \cite{wang-etal-2017-steering,TACL1424}
and interacting fluently with humans \cite{ghazvininejad2018knowledge,zhang2019dialogpt,adiwardana2020humanlike}. 

\subsection{Diverse Decoding}
One major issue with \sts systems is their propensity to select dull, non-committal responses regardless of the input, for which many diverse decoding algorithms have been proposed to tackle this problem \cite{li2015diversity,li2016mutual,vijayakumar2016diverse,cho2016noisy-parallel,kulikov2018importance,kriz-etal-2019-complexity,ippolito-etal-2019-comparison}. 
\citet{li2015diversity}  proposed to use Maximum Mutual Information (MMI) as the objective function in neural dialog models. MMI models use both the forward probability $p(y|x)$ and the backward probability $p(x|y)$ to better capture the contextual relations between the source and target sequences. 
\citet{li2016mutual} introduced a Beam Search diversification heuristic to discourage sequences from sharing common roots, implicitly resulting in diverse sequences. 
\citet{vijayakumar2016diverse} improved upon \citet{li2016mutual} and presented Diverse Beam Search, which formalizes beam search as an optimization problem and augments the objective with a diversity term. 
\citet{cho2016noisy-parallel} introduced Noisy Parallel Approximate Decoding, a method encouraging diversity by adding small amounts of noise to the hidden state of the decoder at each step, instead of manipulating the probabilities outputted from the model.
\citet{kulikov2018importance} attempted to explore larger beam search space by running beam search many times, where the states explored by subsequent beam searches are restricted based on the intermediate states explored by previous iterations.
These works have pushed dialogue models to generate more interesting and diverse responses that are both high-quality and meaningful.

\subsection{Non-Autoregressive Sequence Generation}
Besides diverse responses, another problem for these dialogue generation models is their autoregressive generation strategy that decodes words one-by-one, making it extremely slow to execute on long sentences, especially on conditions where multi-turn dialogue often appears \cite{adiwardana2020humanlike}. One solution is to use non-autoregressive sequence generation methods, which has recently aroused general interest in the community of neural machine translation (NMT) \cite{jiatao2018nat,lee2018iterative,xuezhe2019flowseq,zhiqing2019natcrf,shu2019latentvariable,bao2019pnat}. \citet{jiatao2018nat} proposed to alleviate latency by using fertility during inference in autoregressive Seq2Seq NMT systems, which led to a $\sim$15 times speedup to traditional autoregressive methods, whereas the performance degrades rapidly.  \citet{lee2018iterative,xuezhe2019flowseq,shu2019latentvariable} proposed to use latent variables to model intermediate word alignments between source and target sequence pairs and mitigate the trade-off between decoding speed and performance. \citet{bao2019pnat} pointed out position information is crucial for non-autoregressive models and thus proposed to explicitly model position as latent variables. \citet{zhiqing2019natcrf} incorporated CRF into non-autoregressive models to enhance local dependencies during decoding.
This work is greatly inspired by these advances in non-autoregressive sequence generation.

\section{Background}
\subsection{Autoregressive \sts Models}
An encoder-decoder model \cite{sutskever2014sequence,vaswani2017attention,bahdanau2014neural}
defines the probability of a target sequence $Y= \{y_1, y_2, ..., y_{L_y}\}$, which is a response in the context of dialogue generation, 
given a source sequence $X= \{x_1, x_2, ..., x_{L_x}\}$, where where $L_x$ and $L_y$ are the length of the source and target sentence respectively. 

An autoregressive encoder-decoder model decomposes the distribution over a target sequence $\bm{y}=\{y_1,\cdots,y_{L_y}\}$ into a chain of conditional probabilities:
\begin{equation}
\begin{aligned}
    p_\text{AR}(\bm{y}|\bm{x};\phi)&= \prod_{t=1}^{L_y+1} \log p(y_{t}| y_{0:t-1}, x_{1:L_x}; \theta)   \\
    &=\prod_{t=1}^m \frac{\exp(f(h_{t-1},e_{y_t}))}{\sum_{y'}\exp(f(h_{t-1},e_{y'}))}
    \end{aligned}
\end{equation}
with $y_0$ being the special  $<BOS>$  token and $y_{L_y+1}$ being the special  $<EOS>$ token. 
The probability of generating a token $y_t$ depends on all tokens in the source $X$, and all its previous tokens $y_{0:{t-1}}$ in $Y$. 
The concatenation of 
$X$ and $y_{0:{t-1}}$ is mapped to a representation $h_{t-1}$ using LSTMs \cite{sutskever2014sequence}, CNNs \cite{gehring2017convolutional} or transformers \cite{vaswani2017attention}. 
$e_{y_t}$ denotes the representation for $y_t$. 

During decoding, the algorithm terminates when the $<EOS>$ token is predicted.
At each time step, either a greedy approach or beam search can be adopted for word prediction.
Greedy search selects the token with the largest conditional probability, the embedding of which is then combined with preceding output to predict the token at the next step. 

\subsection{Non-Autoregressive \sts Models}
\subsubsection{Overview}
The autoregressive generation  model has two major drawbacks: it prohibits generating multiple tokens simultaneously, which leads to inefficiency in GPU usage; and 
erroneously generated tokens leads to error accumulation and the performance of beam search deteriorates when exposed to a larger search space \citep{koehn-knowles-2017-six}. 
 Non-autoregressive methods address these two issues by removing the sequential dependencies within the target sentence and generating {\it all} target tokens simultaneously, with the probability giving as follows:   
 \begin{equation}
    p_\text{Non-AR}(\bm{y}|\bm{x};\phi)=\prod_{t=1}^{L_y} p(y_t|\bm{x};\phi)
\end{equation}
Now that each target token $y_t$ only depends on the source sentence $\bm{x}$, the full target sentence can be decoded in parallel, where \verb|argmax| is applied to each token. A vital challenge
that non-autoregressive face is  the {\it inconsistency problem} \citet{jiatao2018nat}, which indicates the decoded sequence contains duplicated or missing tokens. Improving decoding consistency on the target side is thus crucial to Non-AR models.

\section{Model}
\subsection{Overview}
The maximum mutual information (MMI) model, proposed in \citep{li2015diversity}, 
tries to find the response that has  the largest value of mutual information with respect to the context.
The form of MMI is given as follows:\footnote{We refer readers to \citep{li2015diversity}  for how Eq.\ref{eqbayesexpanded} is obtained. }
\begin{equation}
\begin{aligned}
\hat{y} = \argmax_{y} \big\{(1-\lambda)\log p(y|x)+\lambda\log p(x|y) \big\}
\label{eqbayesexpanded}
\end{aligned}
\end{equation}
This weighted MMI objective function can  be viewed as representing a tradeoff between sources given targets (i.e., $p(x|y)$) and targets given sources (i.e., $p(y|x)$). 
Direct decoding from $\log (1-\lambda)p(y|x) + \lambda\log p(x|y)$
is infeasible 
 since the
second part (i.e., $p(x|y)$) requires the completion of target generation before $p(x|y)$ can be  computed. Empirically, an N-best list is first generated given $p(y|x)$, and $p(x|y)$ is then used to rerank the N-best list, which 
inevitably results
in non-globally-optimal solutions.

Here to propose to use Non-AR generation models to handle to non-globally-optimality issue.
The generation of each target word $y_t$ is independent under the non-AR formalization, and 
the forward probability $p(y|x)$ is given as follows:
\begin{equation}
\text{forward\_prob} = \prod_{t=1}^{t=L_y} p(y_t|x)
\label{forward}
\end{equation}

For the backward probability $p(x|y)$, which denotes the probability of generating a source sequence given 
a target sequence,
we  propose to  replace it  with the geometric mean of the probability of generating the source sequence given 
each target token, denoted as follows:
\begin{equation}
\text{backward\_prob} = [\prod_{t=1}^{t=L_y} p(x|y_t) ]^{1/L_y}
\label{backprob}
\end{equation}
We also use the non-AR framework to model the backward probability. 
Based on the independence assumption of non-AR, in which the generations of $x_t$ are independent, 
Eq. \ref{backprob} can be further factorized as follows:
\begin{equation}
\text{backward\_prob} = [\prod_{t=1}^{t=L_y} \prod_{t'=1}^{t'=L_x} p(x_{t'}|y_t) ]^{1/L_y}
\label{back}
\end{equation}
A close look at Equ.\ref{back} shows that it actually mimics  
the 
IBM model  \cite{brown1993mathematics}: 
 $p(x_{t'}|y_t)$ handles the pairwise word alignment between sources and targets. Since position representations are incorporated at both the encoding and decoding stage, 
 Eq.\ref{back} actually mimics IBM model2, where relative positions between source and target words are modeled. 

\begin{figure}
    \centering
    \includegraphics[scale=0.5]{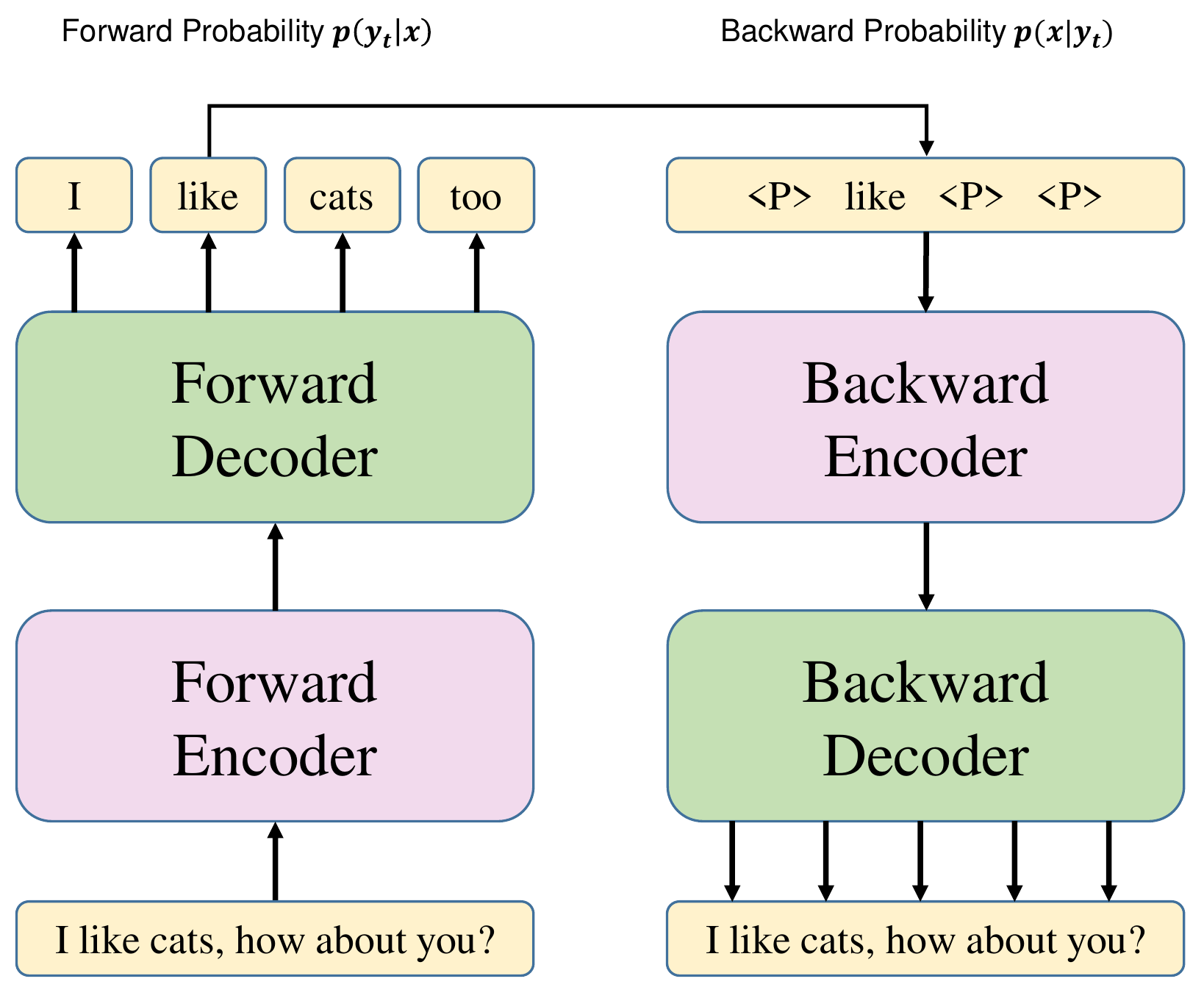}
    \caption{Overview of the non-auto MMI generation model.}
    \label{fig:overview}
\end{figure}

Combining the forward probability in Eq. \ref{forward} and  the backward probability in Eq.\ref{back}, 
the full form of mutual information of 
Eq.\ref{eqbayesexpanded} can be rewritten as follows:
\begin{equation}
\begin{aligned}
L = & (1-\lambda) \sum_{t=1}^{t=L_y}\log p(y_t|x)+ \frac{\lambda}{L_y} \sum_{t=1}^{t=L_y} \sum_{t'=1}^{t'=L_x} \log p(x_{t'}|y_t)  \\
=& \sum_{t=1}^{t=L_y} [ (1-\lambda) \log p(y_t|x) +  \frac{\lambda}{L_y}   \sum_{t'=1}^{t'=L_x} \log p(x_{t'}|y_t)] 
\label{fill}
\end{aligned}
\end{equation}
as can be seen, we are able to factorize the full form of the MMI objective 
with respect to 
 $y_t$ under the framework of non-AR generation. This means that the mutual information between source $x$ and different target words $y_t$  are independent and can be computed in parallel. 
 Also, 
for each token $y_t$, its mutual information with respect to the source $x$ can be readily  computed as soon as $y_t$ is generated, and we do not have to wait until 
the completion of the entire sequence. This naturally resolves the non-globally-optimality issue in the AR generation model. 
Figure~\ref{fig:overview} gives an illustration for the proposed model. 

\subsection{Forward Probability $p(y|x)$}
\label{forward}
We use the non-autoregressive \sts model as the backbone to compute $\prod_t p(y_t|x)$, which consists of two major components: the encoder and the decoder. 
\subsubsection{Encoder}
We use transformers \cite{vaswani2017transformer} as a backbone and use a stack of $N=6$ identical transformer blocks as the encoder. Given the source sequence $\bm{x}=\{x_1,\cdots,x_n\}$, the encoder produces its contextual representations $\bm{H}=\{\bm{h}_1,\cdots,\bm{h}_n\}$ from the last layer of the encoder.

\subsubsection{Decoder}
\paragraph{Target Length} We first need to obtain the length of the target sequence for decoding. We follow previous works \cite{jiatao2018nat,xuezhe2019flowseq,bao2019pnat} to predict the length difference $\Delta m$ between source and target sequences using a classifier with a range of [-20, 20]. This is accomplished by  max-pooling the source embeddings into a single vector, running this through a linear layer followed by a softmax operation, as follows:
\begin{equation}
    p(\Delta m|\bm{x})=\text{softmax}(W_{p}(\text{maxpool}(\bm{H}))+b_p)
\end{equation}

\paragraph{Decoder Structure}The decoder also consists of $N=6$ identical transformer blocks. 
The $i$-th position of the input $\bm{d}_i$ to the decoder is 
 the $\text{round}(n*(i/m))$-th input's contextual representation $\bm{h}_{\text{round}(n*(i/m))}$ copied from the encoder, which is equivalent to scanning the source inputs from left to right and leads to a deterministic decoding process given the predicted target length. 
Both absolute and relative positional embeddings are incorporated. 
For {relative} position information, we follow \citet{shaw2018relativepos} which produces a different learned embedding according to the offset between the ``key'' and ``query'' in the self-attention mechanism with a clipping distance $k$ (we set $k=4$) for relative positions.
For absolute positional embeddings, we follow \citet{radford2019gpt2} and used a learnable positional embedding $\bm{p}_t$ for position $t$.

\paragraph{Attention over Vocabulary}
 Layer-wise attention over vocabulary is incorporated into each decoding layer to make the model aware of which token is to be generated regarding each position. 
More concretely, we use $\bm{Z}^{(i)}(1\le i\le 6)$ to denote the contextual representations for the $i$-th decoder layer , and $\bm{Z}^{(0)}=\{\bm{d}_1,\cdots,\bm{d}_m\}$
to denote 
 the input to the decoder. 
The intermediate token attention representation $\bm{a}^{(i)}_j$ of position $j (1\le j\le m)$ in the $i$-th decoder layer is thus given by:
\begin{equation}
    \bm{a}^{(i)}_j=\text{softmax}(\bm{z}^{(i)}_j\cdot\bm{W}^\mathrm{T})\cdot\bm{W}
\end{equation}
where $\bm{W}$ is the representation matrix of the token vocabulary. By doing so, each position is able to know which token is about to be decoded at the current position. 
The input to the next layer 
 $\bm{Z}^{(i+1)}$ is the concatenation of the contextual representations and the intermediate token representations $[\bm{Z}^{(i)};\bm{A}^{(i)}]$ . 
\paragraph{softmax}
For each position $t$,  $p(y_t|x)$ is computed by outputting the representation for that position to a softmax function. 
\subsection{Backward Probability $p(x|y)$}
We use the non-AR model to obtain $p(x|y_t)$. 
\subsubsection{Encoder}
The encoder for $p(x|y_t)$ is again a stack of $N=6$ identical transformer blocks.
The input to the encoder is a text sequence  with  length being $L_y$, which is identical to the length of the target.
The $t$-th position of the input sequence is the word $y_t$, with the rest being the place-holding dummy token. 
For each posiition, 
the embedding for the 
absolute position  and 
the embedding for the
relative position  are appended. 
\subsubsection{Decoder}
The decoder for the backward probability is the same as that of the forward probability, with the only difference being changing  target $y$ to source $x$.

\subsection{Decoding from Mutual Information}

The most commonly used decoding strategy for non-AR generation is the noisy parallel decoding strategy (NPD for short) proposed in \citet{jiatao2018nat}:
a number of sequence candidates are first generated by the non-AR generation, then an AR  \sts model is used to select the candidate that has the largest value of probability 
output from the AR model. 
Since this NPD strategy is used for the MLE objective which only concerns about the forward probability, we need to tailor it to the MMI objective. 
Specifically, we first generate N-best sequences based on the score  of non-AR MMI function, computed from Eq.\ref{fill}.
The final selected response is the sequence with highest AR MMI score, which is computed based on two  AR \sts models, one to model the forward probability and the other to model 
the backward probability.

\section{Experiments}
\subsection{Datasets}
We use the OpenSubtitles dataset for evaluation. 
It's a widely used  open-domain dataset, which contains roughly 60M-70M scripted lines spoken by movie characters.  
It
has been used in a broad range of recent
work on data-driven conversation
This dataset does not specify which character speaks each subtitle line, which prevents us from inferring speaker turns. 
Following \citep{vinyals2015neural,li2015diversity}, we make an assumption that each line of subtitle constitutes a full speaker turn.
 Although this assumption is often violated, prior work has successfully trained
and evaluated neural conversation models using
this corpus. 
In our experiments we used a preprocessed version of this dataset distributed by \citet{li2015diversity}.\footnote{\url{
http://nlp.stanford.edu/data/OpenSubData.tar}}

The noisy nature of the OpenSubtitle 
dataset renders it unreliable for evaluation purposes. 
We thus follow \citet{li2015diversity} 
to use 
 data from the Internet Movie Script Database (IMSDB)\footnote{ \url{http://www.imsdb.com/}} for evaluation.
 The IMSDB dataset 
  explicitly identifies which character speaks each line of the script. 
We followed protocols in \cite{li2015diversity} 
and randomly selected two subsets as development and test datasets, each containing 2,000 pairs, with source and target length restricted to the range of [6,18]. 


\subsection{Baselines}
Our baselines include the AR generation models (using or not using MMI) based on transformers \cite{vaswani2017attention}, with the number of encoder and decoder blocks set to 6. 
For the standard AR model, the value of beam size is set to 10 for decoding, and the sequence with the largest value of $p(y|x)$ is selected.  
For AR+MMI, we followed \citet{li2015diversity}, and first use $p(y|x)$ to generate an N-best list with beam-size 10. Then $p(x|y)$ is used to rerank the N-best list.
$\lambda$ is treated as the hyper-parameter to be tuned on the dev set. 

We also implement two variant of the AR+MMI model: (1) AR+MMI+diverse \cite{li2016simple}, which uses a diverse decoding model to generate the N-best list and uses the backward probability to rerank the diverse N-best list. 
The diverse decoding model  adds an
additional term  to penalize siblings in beam search—expansions of
the same parent node in the search— thus favoring
choosing hypotheses from diverse parents; and (2) AR+MMI+RL \cite{li2017learning}, which incorporates
 the critic that estimates further backward probability into decoding. 
\subsection{Training Details}
All experiments were run using 64 Nvidia V100 GPUs with mini-batches of approximately 100K
tokens. 
We use the same hyper-parameters for all experiments, i.e., word representations of size 1024, feed-forward
layers with inner dimension 4096. Dropout rate is set to 0.2 and the number of attention heads is set to 16.
Models are optimized with Adam \cite{kingma2014adam} using $\beta_1=0.9$, $\beta_2 = 0.98$,
$\epsilon = 1e − 8$. 
Differentiable scheduled sampling \citet{goyal2017differentiable} is used to mitigate the exposure bias issue. 
We train models with 16-bit floating point
operations. 
The backward model and the forward model are jointly trained with word embeddings shared. 
\begin{table*}[!ht]
\center
\begin{tabular}{ccccccc}
Model&BLEU&distinct-1&distinct-2&Avg.length&Stopword & adv succ \\\hline
Human& - & 16.8\% & 58.1\% & 14.2 & 69.8\% &  \\ \hline
AR&1.64 & 3.7\% & 9.5\% & 6.4 &  82.3\% &2.7\%   \\
AR+MMI &  2.10 & 10.6\% & 20.5\% & 7.2& 76.4\%&6.3\%  \\ 
AR+MMI+diverse  & 2.16& 16.0\% & 27.3\%& 7.5 & 72.1\% &  6.4\%\\
 AR+MMI+RL & 2.34 & 13.7\% & 25.2\% & 7.3 & 73.0\%& 8.0\% \\ \hline
NonAR& 1.54 & 8.9\% &  14.6\% &7.1&   77.9\%& 2.4\% \\
NonAR+MMI & 2.68 & 15.9\% &  27.0\%& 7.4 & 71.9\% & 9.2\% \\\hline
\end{tabular}
\caption{Automatic Metrics Evaluation for Different Models. }
\label{autometric}
\end{table*}

\subsection{Automatic Evaluation}
For automatic evaluation, we report the results of the following metrics:
\begin{itemize}
\item the  BLEU score  following previous work.
It should be noted that BLEU is not generally accepted \cite{liu2016not} to match human evaluation in generation tasks since there are distinct ways to reply to an input.
\item
 {\it distinct-1} and {\it distinct-2} \cite{li2015diversity}: calculating the number of distinct unigrams and bigrams in generated responses scaled by total number of generated unigrams and bigrams.
\item  Avg.length: the average length of the generated response. 
\item Stopword\%: the percentage of stop-words\footnote{\url{The combination of stopwords in https://www.ranks.
nl/stopwords and punctuations.}} of the responses generated by each model. 
\item Adversarial Success: 
the adversarial evaluation strategy proposed by \citet{kannan2017adversarial,li2017adversarial}. Adversarial evaluation 
trains a discriminator (or evaluator) function to labels dialogues as
machine-generated (negative) or human-generated
(positive). Positive examples are taken from
training dialogues, while negative examples are
decoded using generative models from a model.
Adversarial success is the percentage of the generated responses that can fool the evaluator to believe that it is human-generated. 
We refer readers to \citet{li2017adversarial} for more
details about the adversarial evaluation. 
\end{itemize}
Results are shown in Table \ref{autometric}. 
When comparing AR with AR+MMI, AR+MMI significantly outperforms AR across all metrics, which is in line with previous findings \cite{li2015diversity}.  
For the variants of AR+MMI, 
AR+MMI+diverse generates a more diverse N-best list for reranking, and thus outperforms AR+MMI;
AR+MMI+RL uses lookahead strategy to estimate future backward probability, and thus outperforms AR+MMI as well. 
It's hard to tell which model performs better, AR or non-AR: AR performs better than non-AR for  BLEU and adversarial success, but worse for the other metrics. This means comparing with AR model, non-AR model tends to generate more diverse responses, but might be less coherent. 
Because of the ability to handle the non-local-optimality issue, Non-AR+MMI  consistently outperforms AR+MMI by a large margin across all evaluation metrics. 
When comparing non-AR with 
AR+MMI+diverse, non-AR has relatively lower {\it distinct} score, but significantly higher scores BLEU  and adversarial success. 
This is because the diverse decoding strategy in 
AR sacrifices language model probability for diversity, and thus harms the BLEU score but promotes the diversity score. 
NonAR+MMI outperforms AR+MMI+RL across all metrics. 

\subsection{Examples}
 \begin{table*}[t]
 \centering
\small
 \begin{tabular}{p{5cm} p{5cm} p{5cm} }\hline
  {\bf Input}& AR+MMI & NonAR+MMI  \\\hline
  it feels like i must have been asleep for weeks. & i ’m sorry that i can 't help you.  & Yeah, it has been long.  \\
  where does she work? & i don 't know where she works. & she has work to do at the library. \\
 who is in charge? &    I don 't know sir. & i 'm the captain in charge! \\
 I am off all week next week. & what time you going? & then who will be in charge ? \\
 why can 't you just believe us? &  I mean , it ’s not like we ’re gonna be together forever . & because you are not a good person. \\
  can 't you see how they ‘re exploiting you? &  i ’m not a human being . & it 's not that simple. \\
  I mean, we’d have to talk to him.  & I mean, he 's a good guy. & let 's do it. \\
  i 'm sorry to detain you for so long. & i 'm sorry i can 't. & that 's fine since i 'm not leaving until tomorrow.   \\
  do you have any idea what caused the explosion ? & i have no idea. & the explosion was caused by the failure of the motor. \\ \hline
  \hline
  \end{tabular}
  \caption{Response generation: Sample responses using the diversity-promoting beam search and vanilla beam search.}
  \label{sample-response}
 \end{table*}

\subsection{Qualitative Evaluation}
 We employed crowdsourced judges to provide evaluations for a random
sample of 1000 items from the test set. 
Following protocols in \citet{baheti2018generating}, 
we assigned each output  to a human judge, who were asked to score every model response 
on a 5-point scale (Strongly Agree, Agree, Unsure,
Disagree, Strongly Disagree)
 on 2 categories: 1) Coherence
- is the response coherent to the given source?
and 2) Content Richness - does the response add
new information to the conversation?
Ratings were later
collapsed to 3 categories (Agree, Unsure, Disagree). 
\begin{table}
\small
\begin{tabular}{cccc}\\\hline
Model  & disagr (\%) &un(\%) &agr(\%) \\\hline
\multicolumn{4}{c}{Coherence} \\\hline
Human & 17.4 & 20.8 & 61.8 \\
AR &  28.6 & 29.5 & 41.9 \\
AR+MMI & 25.3 & 27.9 & 46.8 \\
AR+MMI+diverse & 24.8 & 27.8 & 47.4 \\
AR+MMI+RL & 24.1 & 26.5 & 49.4 \\
nonAR & 29.9 & 28.7 & 41.4 \\
nonAR+MMI & 23.1 & 24.0 & 52.9 \\\hline
\multicolumn{4}{c}{Content Richness} \\\hline
Human & 14.0 & 16.6 & 69.4 \\
AR & 38.2 & 30.4 & 31.4 \\
AR+MMI & 30.6 & 26.2 & 43.2 \\ 
AR+MMI+diverse & 23.9 & 21.3 & 54.8 \\
AR+MMI+RL & 26.4 & 24.9 & 48.7 \\
NonAR & 31.4 & 25.0 & 44.6 \\
NonAR+MMI& 24.2 & 20.5 & 55.3 \\\hline
\end{tabular}
\caption{ Human judgments for Coherence and Content Richeness of the different models.}
\label{human}
\end{table}
\begin{table*}[t]
    \small
    \centering
    \scalebox{0.85}{
      \setlength{\tabcolsep}{4pt}
      \begin{tabular}{l@{\hspace{0.3cm}}c@{\hspace{0.25cm}}c@{\hspace{0.25cm}}c@{\hspace{0.25cm}}c@{\hspace{0.25cm}}c@{\hspace{0.25cm}}
      c@{\hspace{0.25cm}}c@{\hspace{0.25cm}}c@{\hspace{0.25cm}}
      }
      \toprule
       & \multicolumn{1}{c}{\bf  WMT14 En$\rightarrow$De}& \multicolumn{1}{c}{\bf WMT14  De$\rightarrow$En}& \multicolumn{1}{c}{\bf WMT16 Ro$\rightarrow$En}
            \\\hline
      NAT \citep{jiatao2018nat} & 17.69  & 20.62 & 29.79    \\
      iNAT \citep{lee2018iterative} & 21.54  & 25.43 & 29.32    \\
      FlowSeq-large (raw data) \citep{xuezhe2019flowseq}  & 20.85  & 25.40 & 29.86 &  \\\hline
      NAT (our implementation) &  22.32 & 24.83 & 29.93\\
      NAT +MMI & 23.80 & 26.05& 30.50  \\
      & (+1.48)  &(+1.22) &(+0.57) \\\hline
            \bottomrule\hline
      
      \end{tabular}
      }
    \caption{The performances of NonAR+MMI methods on WMT14 En$\leftrightarrow$De and WMT16 Ro$\rightarrow$En. Results from \citet{jiatao2018nat,lee2018iterative,xuezhe2019flowseq} are copied from original papers for reference purposes. }
    \label{MT}
    \vskip -0.15in
    \end{table*}

The results for plausibility and content
richness of different models are presented in Table \ref{human}.
For dialogue coherence, the trend is that NonAR+MMI is better than AR+MMI, followed by AR and Non-AR. AR is slightly better than Non-AR.  
For Content Richness, the proposed NonAR+MMI is significantly better than AR+MMI, and the gap is greater than dialogue coherence.
This is because the N-best list generated by the AR model tends to be dull and generic, and the reranking model in AR+MMI can help alleviate but cannot fully address this issue.
The output from the AR+MMI model is thus by far less diverse than nonAR+MMI, which obtains the MMI score for each generated token.

To verify the statistical significance of the reported results, we performed a pairwise bootstrap test  \cite{johnson2001introduction,berg2012empirical}
to compare the difference between percentage of responses that were labeled as yes. 
We computed p-values for non-AR+MMI vs AR+MMI and non-AR vs AR.
Regarding non-AR vs AR, we did not find a significant difference (p-value = 0.18) for coherence, 
but a significant difference for content richness (p-value $<$ 0.01).
For non-AR+MMI vs AR+MMI and  AR+MMI+RL, 
we find a significant difference for both 
coherence (p-value $<$ 0.01) and content richness (p-value $<$ 0.01). 
For non-AR+MMI vs AR+MMI+RL, the difference for coherence is significant (p-value $<$ 0.01), 
but content richness is insignificant (p-value=0.25). 

\subsection{Sample Responses}
Sample responses are presented in Table \ref{sample-response}. 
As  can be seen, the nonAR+MMI tends to generate more diverse and content-rich responses. 
It is also interesting to see that  responses from the AR+MMI model mostly start with the word ``{\it I} ". 
This is because of the fact that the N-best list from the AR model lacks for diversity. The prefixes of the responses are mostly the same and the 
reranking process can only affect suffixes. 
On the contrary, for  nonAR+MMI, MMI reranking is performed once a token is generated, and does not wait for the completion of the whole target sequence, leading to more diverse and appropriate responses. 
 \subsection{Results on Machine Translation}
 Mutual information has been found to improve machine translation, both in the context of NMT models \cite{li2016mutual} and
 phrase-based MT models \cite{och2002discriminative,shen2010string}. 
 It would be interesting to see whether the proposed model can also help non-AR NMT as well. 
 We  evaluate the proposed method on the three widely used machine translation benchmark tasks (three  datasets):
WMT2014 De$\rightarrow$En (4.5M sentence pairs), WMT2014 En$\rightarrow$De, WMT2016 Ro$\rightarrow$En (610K sentence pairs) and IWSLT2014 De$\rightarrow$En (150K sentence pairs). We use the Transformer \citep{vaswani2017transformer} as a backbone.  
{\bf Knowledge Distillation} is applied for all models. 
Since building SOTA non-AR MT models is out of the scope of this paper, we used the commonly used NonAR structure described in Section \ref{forward} as the backbone. 
Results are shown in Table \ref{MT}. As can be seen, the incorporation of MMI model  significantly improves MT performances. 
 This shows that the proposed model has potentials to benefit a wide range of generation tasks. 
 
\section{Conclusion}
In this paper, we propose to use non-autoregressive (non-AR) generation to address the  non-global optimality issue for MMI in neural dialog generation. Target tokens are generated independently in non-AR generation.
$p(x|y)$ for each target word can thus be computed as soon as it s generated, and does not have to wait for the completion of the whole sequence.
This naturally resolves the non-global optimal issue in decoding. 
Experimental results demonstrate that the proposed strategy produces more diverse, coherent, and appropriate responses, yielding substantive gains in BLEU scores  and in human evaluations.

\bibliography{nmt}
\bibliographystyle{acl_natbib}

\end{document}